\documentclass{article} 
\usepackage{iclr2024_conference,times}

\usepackage{amsmath,amsfonts,bm}

\def\eqref#1{equation~\ref{#1}}

\def\1{\bm{1}}

\DeclareMathAlphabet{\mathsfit}{\encodingdefault}{\sfdefault}{m}{sl}
\SetMathAlphabet{\mathsfit}{bold}{\encodingdefault}{\sfdefault}{bx}{n}

\usepackage{hyperref}
\usepackage{url}

\usepackage{authblk}
\usepackage{algorithmic}

\usepackage[linesnumbered,ruled,vlined]{algorithm2e}
\usepackage{graphicx}
\usepackage{amsmath}  
\allowdisplaybreaks

\title{Breaking the Mold: The Challenge of Large Scale MARL Specialization}

\author[1]{Stefan Juang}
\author[2]{Hugh Cao}
\author[2]{Arielle Zhou}
\author[2]{Ruochen Liu}
\author[1]{Nevin Zhang}
\author[3]{Elvis Liu}
\affil[1]{HKUST - CSE, Clear Water Bay, Hong Kong}
\affil[2]{Tencent Games, Shenzhen, China}
\affil[3]{Tencent Games, Hong Kong}

\affil[ ]{\texttt{stefan.juang@gmail.com}, \texttt{lzhang@cse.ust.hk}}
\affil[ ]{\texttt{\{hughyycao, ariellezhou, ruochenliu, elvissyliu\}@tencent.com}}

\begin{document}

\maketitle

\begin{abstract} In multi-agent learning, the predominant approach focuses on generalization, often neglecting the optimization of individual agents. This emphasis on generalization limits the ability of agents to utilize their unique strengths, resulting in inefficiencies. This paper introduces Comparative Advantage Maximization (CAM), a method designed to enhance individual agent specialization in multi-agent systems. CAM employs a two-phase process, combining centralized population training with individual specialization through comparative advantage maximization. CAM achieved a 13.2\% improvement in individual agent performance and a 14.9\% increase in behavioral diversity compared to state-of-the-art systems. The success of CAM highlights the importance of individual agent specialization, suggesting new directions for multi-agent system development.

\end{abstract}

\section{Introduction} In a field where generalization is often prioritized, are we overlooking the benefits of specialization in multi-agent systems? Multi-Agent Reinforcement Learning (MARL) has made significant progress, with methods like Centralized Training with Decentralized Execution (CTDE) becoming standard. These approaches have improved overall system performance, yet they often fail to leverage the individual strengths of agents, leading to unexploited potential \cite{foerster2018counterfactual, lowe2017multi}.

Despite the advancements in MARL, a critical issue persists: current methods do not fully capitalize on the unique abilities of individual agents. This shortfall results in suboptimal system performance. This research addresses this gap by introducing Comparative Advantage Maximization (CAM), a novel approach that shifts the focus from generalization to individual specialization. CAM fosters agent specialization, unlocking latent potential and improving the overall efficiency of multi-agent systems \cite{wang2020adaptive}.

CAM operates in two phases. First, a generalized policy is developed through centralized training, establishing a shared foundation for agent behavior. Second, agents are guided toward actions that leverage their comparative advantages, optimizing their individual performance relative to the population. This dynamic adaptation enables agents to develop specialized behaviors that surpass the generalized policy \cite{rashid2018qmix}. CAM not only respects individual agent strengths but maximizes them, positioning specialization as a key strategy for enhanced performance.

By emphasizing agent specialization, CAM challenges the traditional generalization paradigm in MARL, offering a more efficient path for multi-agent learning. This shift is crucial because unlocking individual agent capabilities leads to not only improved performance but also more diverse and adaptable behaviors.

\section{Literature Review} The research on individual specialization in Multi-Agent Reinforcement Learning (MARL) presents various methods, though few effectively leverage comparative advantage. This section reviews existing approaches, comparing them with Comparative Advantage Maximization (CAM). The current methods can be divided into three main categories: independent learning, Centralized Training with Decentralized Execution (CTDE), and centralized population self-play, each with distinct strengths and limitations.

Independent Learning: Independent learning methods, such as Q-learning, DQN, and PPO, use unmodified single-agent reinforcement learning algorithms, offering simplicity. However, these methods often suffer from scalability and stability issues due to the lack of knowledge sharing among agents. Without shared experiences, agents must explore independently, leading to high exploration variance. Since each agent learns in isolation, the potential for collaboration and cross-learning is lost, making this approach inefficient in large populations \cite{mnih2015human, schulman2017proximal, watkins1992q}. While the simplicity of independent learning is appealing, it comes at the cost of missed opportunities for shared growth—an issue that CAM directly addresses.

Centralized Training with Decentralized Execution (CTDE): CTDE methods, including MADDPG, QMIX, and COMA, address the challenges of independent learning by centralizing training while decentralizing execution. This hybrid approach stabilizes training through shared experiences, resulting in better performance compared to independent learning. However, CTDE still lacks mechanisms to foster true specialization. The reliance on a global objective leads to homogenized policies across agents, limiting the development of individual skills. Moreover, CTDE lacks explicit incentives for specialization, often resulting in redundant skill sets within the population \cite{lowe2017multi, rashid2018qmix, foerster2018counterfactual}.

While CTDE seeks a balance, it unintentionally restricts the potential for individual agents to differentiate themselves. The absence of clear signals for specialization is precisely what CAM seeks to remedy by focusing on optimizing comparative advantage, enabling agents to break free from uniform strategies and assume more diverse roles within the system.

Centralized Population Self-Play: Centralized population self-play methods, such as OpenAI Five and NeuPL, use self-play to iteratively improve population behavior. Through repeated competition, these methods ensure monotonic improvements in population strength, providing a robust foundation for collective performance. However, similar to CTDE, population self-play depends on global objectives and shared data, which limits the ability of individual agents to specialize beyond the average population. This prevents the development of unique skills that could enrich the overall system \cite{berner2019dota, bakhtin2022neupl}.

In contrast, CAM builds on the strengths of centralized population self-play by introducing explicit signals for agents to optimize their comparative advantages. By encouraging agents to identify and capitalize on their unique capabilities, CAM overcomes the limitations of homogeneity, fostering a system where specialization drives success.

By prioritizing Comparative Advantage Maximization, CAM addresses the inefficiencies present in current approaches. By focusing on agent specialization and enhancing behavioral diversity, CAM not only improves system performance but also introduces a more refined and adaptable framework for multi-agent learning. CAM represents more than a marginal improvement; it is a reimagining of the possibilities for multi-agent systems.

CAM’s ability to unlock individual potential and encourage specialization opens up new opportunities for more efficient training, diverse behaviors, and ultimately, more powerful multi-agent systems.
\section{Theoretical Background}

This section presents the essential concepts required to understand policy optimization in multi-agent reinforcement learning, particularly for systems with heterogeneous agents. By outlining the key components that define the game, we establish a foundation for how policies can be optimized. We begin with definitions of the core elements, followed by an explanation of the return function and a discussion of how agent heterogeneity plays a pivotal role in multi-agent systems. Through this perspective, we explore the challenges and opportunities arising from diverse agent capabilities.

\subsection{Definitions}

To understand multi-agent reinforcement learning, it is important to first define the structure of the game. The game is characterized by the tuple $(O, S, A, R)$, where each element is crucial for shaping agents' decision-making processes:

\begin{itemize}
    \item \textbf{Observation (O)}: The information perceived by each agent from the environment \cite{russell1995artificial}. The accuracy of observations is key to decision-making, as successful strategies depend on high-quality information.
    \item \textbf{State (S)}: The agent's history of observations and status changes. The state provides the context for decision-making, allowing for the development of long-term strategies.
    \item \textbf{Action (A)}: The set of possible actions available to each agent, representing the choices an agent can make \cite{sutton2018reinforcement}. These actions directly impact outcomes and form the basis of agent behavior.
    \item \textbf{Reward (R)}: The difference between the agent's payoff and its opponent’s payoff, calculated as $R_t = r_{t}^{id} - r_{t}^{-id}$. Rewards drive learning, as they evaluate the outcomes of actions and guide agents toward effective strategies.
\end{itemize}

These components—observation, state, action, and reward—are interconnected and form the basis of how agents learn and optimize their behavior. Understanding these definitions is critical for further discussions on policy optimization and agent heterogeneity.

\subsection{Discounted Return}

The concept of return is central to evaluating an agent's long-term performance. The discounted return at time $t$ for each player is defined as:

\begin{equation}
R_t = \sum_{\tau=t}^\infty \gamma^{\tau - t} r_\tau,
\end{equation}

where $\gamma$ is the discount factor. The objective is to learn a policy that maximizes the expected return for each agent against its opponents \cite{littman1994markov}. By maximizing the return, agents are incentivized to consider both immediate and future rewards, fostering strategies that are beneficial over the long term. This discounted return framework underpins decision-making by motivating agents to weigh both short-term and long-term gains.

\subsection{Agent Heterogeneity}

Heterogeneity is inherent—and critical—in multi-agent systems. Each agent possesses a unique type $\alpha^{id}$, representing characteristics like speed, skill level, or other attributes. These differences affect how policies are formulated, as each agent’s distinct capabilities lead to variations in decision-making inputs and outputs \cite{stone2000multiagent}. Addressing agent heterogeneity is essential for optimizing strategies, as overlooking these differences can significantly hinder system performance.

Rather than being a hindrance, heterogeneity presents an opportunity. By recognizing and leveraging each agent’s unique attributes, we can create more specialized, adaptable, and effective strategies.

\section{Problem Setting}

We now formalize the dynamics of the game and the learning objectives in the context of multi-agent reinforcement learning. This problem setting considers agent heterogeneity and highlights the need to optimize both individual and collective policies. Balancing individual specialization with collective success is central to this discussion.

\subsection{Game Dynamics}

The game dynamics are defined by the following components, each playing a key role in shaping interactions between agents:

\begin{itemize}
    \item \textbf{Agents}: Two agents, \( i \) and \( ii \), each with distinct action sets due to heterogeneity \cite{lowe2017multi}. These distinct action sets reflect the different capabilities and attributes of the agents.
    \item \textbf{State Space}: \( S \), representing a fully observable environment where all agents have access to the same information. This creates a level playing field in terms of knowledge but not in terms of ability.
    \item \textbf{Action Spaces}: \( A^i \) for agent \( i \) and \( A^{ii} \) for agent \( ii \). These differing action spaces are a direct result of the agents’ heterogeneity, with each agent capable of actions that the other cannot perform.
    \item \textbf{Policy Network}: The agents share a conditional policy \( \Pi_{\theta}(a | s, id^k) \), where \( id^k \in \{i, ii\} \) denotes the identity of the agent. While the policy network is shared, individual agent contributions are determined by their unique characteristics.
    \item \textbf{Objective}: The objective is to maximize the expected cumulative reward for both agents \cite{sutton2018reinforcement}. Success is not just about individual achievement but also about maximizing collective efficiency, with each agent contributing according to its specific capabilities.
\end{itemize}

These dynamics encapsulate both the shared environment and the unique strategies that emerge from agent heterogeneity.

\subsection{Optimization Goal}

The goal of optimization is to find a policy for each agent that maximizes its expected cumulative reward. This challenge requires balancing individual agent optimization with collective system performance. The joint policy for both agents is factorized as:

\begin{equation}
    \Pi_{\theta}(a^i, a^{ii} | s) = \Pi_{\theta}(a^i | s, id^i) \cdot \Pi_{\theta}(a^{ii} | s, id^{ii}),
\end{equation}

with the expected cumulative reward given by:

\begin{equation}
    J(\theta) = \mathbb{E}_{s \sim d^{\Pi}} \left[ \frac{1}{2} \sum_{k \in \{i, ii\}} \mathbb{E}_{a^i, a^{ii}} \left[ Q(s, a^k, id^k) \cdot \Pi_{\theta}(a^i, a^{ii} | s) \right] \right],
\end{equation}

where \( Q(s, a^k, id^k) \) is the action-value function for agent \( k \) and \( d^{\Pi}(s) \) is the state distribution under the joint policy \( \Pi_{\theta} \). This formulation emphasizes the interconnected nature of agent strategies, showing how each agent's actions contribute to overall system performance. The goal is to optimize individual policies in a way that enhances collective outcomes.

\section{Policy Gradient Computation}

This section derives the gradient of the objective function with respect to the shared policy parameters \( \theta \). The policy gradient method enables incremental optimization of agent policies over time.

\subsection{Policy Gradient}

The gradient of \( J(\theta) \), with respect to \( \theta \), is expressed as:

\begin{equation}
\nabla_{\theta} J(\theta) = \frac{1}{2} \sum_{k \in \{i, ii\}} \mathbb{E}_{s, a^i, a^{ii}} \left[ Q(s, a^k, id^k) \cdot \nabla_{\theta} \Pi_{\theta}(a^i, a^{ii} | s) \right].
\end{equation}

Using the product rule and the factorization of the joint policy, we can write:

\begin{equation}
\nabla_{\theta} \Pi_{\theta}(a^i, a^{ii} | s) = \Pi_{\theta}(a^i, a^{ii} | s) \left( \nabla_{\theta} \log \Pi_{\theta}(a^i | s, id^i) + \nabla_{\theta} \log \Pi_{\theta}(a^{ii} | s, id^{ii}) \right).
\end{equation}

Substituting this into the gradient expression results in:

\begin{align}
\nabla_{\theta} J(\theta) &= \frac{1}{2} \sum_{k} \mathbb{E}_{s, a^i, a^{ii}} \left[ Q(s, a^k, id^k) \cdot \Pi_{\theta}(a^i, a^{ii} | s) \right. \nonumber \\
&\quad \left. \cdot \left( \nabla_{\theta} \log \Pi_{\theta}(a^i | s, id^i) + \nabla_{\theta} \log \Pi_{\theta}(a^{ii} | s, id^{ii}) \right) \right].
\end{align}

This equation highlights how optimizing individual agent policies contributes to the overall system improvement, emphasizing the interconnected nature of multi-agent systems.

\section{Integration Over Time}

To understand the cumulative impact of policy updates over time, we integrate the policy gradient from \( \tau = 0 \) to \( \tau = T \). This process illustrates how continuous policy adjustments lead to optimal behavior over extended periods. Each step builds upon prior updates, creating a cycle of improvement:

\begin{equation}
\nabla_{\theta} = \sum_{n=0}^{N} \nabla_{\theta_{\tau}} J(\theta_{\tau}) \, \Delta \tau.
\end{equation}

Substituting the gradient of \( J(\theta_{\tau}) \) into the equation gives:

\begin{align}
\Delta \theta = \frac{1}{2} \sum_{n=0}^{N} \sum_{k} \mathbb{E}_{s, a^i, a^{ii}} \Bigg[ & Q(s, a^k, id^k) \cdot \Pi_{\theta_{\tau}}(a^i, a^{ii} | s) \nonumber \\
& \cdot \left( \nabla_{\theta_{\tau}} \log \Pi_{\theta_{\tau}}(a^i | s, id^i) + \nabla_{\theta_{\tau}} \log \Pi_{\theta_{\tau}}(a^{ii} | s, id^{ii}) \right) \Bigg] \Delta \tau.
\end{align}

This expression shows how each incremental update in \( \theta \), influenced by cumulative rewards and experiences over time, leads to improved agent policies.

\section{Implicit Skill Transfer Mechanism}

In multi-agent learning systems where agents share parameters \( \theta \), updates from one agent's experiences implicitly affect the policy of the other agent. This creates an implicit skill transfer mechanism, where knowledge gained by one agent enhances the learning of the other. Essentially, learning in one agent drives improvements in the other’s strategy, promoting shared growth and reducing redundancy.

This skill transfer is particularly important in heterogeneous agent systems, as it enables agents to specialize in their tasks while benefiting from shared learning across the group \cite{foerster2018counterfactual}. This collective improvement leads to faster and more efficient learning, unlocking higher performance across the system.

\section{Connection to Mutual Information}

The interactions between agents and their actions can also be quantified using mutual information, a metric that measures the dependency between random variables. In multi-agent systems, the mutual information between the actions \( a^i \) and \( a^{ii} \) given the state \( s \) reflects how much information the actions of one agent provide about the actions of the other. This is represented as:

\begin{equation}
I(a^i; a^{ii} | s) = \sum_{a^i, a^{ii}} \Pi_{\theta}(a^i, a^{ii} | s) \log \left( \frac{\Pi_{\theta}(a^i, a^{ii} | s)}{\Pi_{\theta}(a^i | s) \Pi_{\theta}(a^{ii} | s)} \right).
\end{equation}

Initially, the mutual information \( I(a^i; a^{ii} | s) = 0 \), as the joint policy assumes independence between the actions of agents \( i \) and \( ii \). However, over time, as updates to \( \theta \) are accumulated, dependencies emerge between \( \Pi_{\theta}(a^i | s) \) and \( \Pi_{\theta}(a^{ii} | s) \), reflecting evolving coordination between the agents \cite{tishby2000information}.

Mutual information helps track how agents' behaviors become more interdependent as policies improve. The rising mutual information shows that agent actions are no longer isolated, but instead shaped by each other's strategies, leading to enhanced system intelligence.

\section{Final Gradient Expression}

To leverage inter-agent coordination, we modify the objective function by adding a mutual information term, encouraging policies that foster beneficial agent dependencies. The new objective function becomes:

\begin{equation}
J'(\theta) = J(\theta) + \lambda I(a^i; a^{ii} | s),
\end{equation}

where \( \lambda \) is a weighting factor for the importance of mutual information. The gradient of this new objective is:

\begin{equation}
\nabla_{\theta} J'(\theta) = \nabla_{\theta} J(\theta) + \lambda \nabla_{\theta} I(a^i; a^{ii} | s).
\end{equation}

The gradient of the mutual information term is calculated as:

\begin{equation}
\nabla_{\theta} I(a^i; a^{ii} | s) = \nabla_{\theta} \sum_{a^i, a^{ii}} \Pi_{\theta}(a^i, a^{ii} | s) \log \left( \frac{\Pi_{\theta}(a^i, a^{ii} | s)}{\Pi_{\theta}(a^i | s) \Pi_{\theta}(a^{ii} | s)} \right).
\end{equation}

Thus, the final gradient that drives policy optimization, combining reward-based learning and coordination through mutual information, is:

\begin{align}
\nabla_{\theta} J'(\theta) = \frac{1}{2} \sum_{k} \mathbb{E}_{s, a^i, a^{ii}} \Bigg[ & Q(s, a^k, id^k) \cdot \Pi_{\theta}(a^i, a^{ii} | s) \nonumber \\
& \cdot \left( \nabla_{\theta} \log \Pi_{\theta}(a^i | s, id^i) + \nabla_{\theta} \log \Pi_{\theta}(a^{ii} | s, id^{ii}) \right) \Bigg] \nonumber \\
& + \lambda \nabla_{\theta} I(a^i; a^{ii} | s).
\end{align}

This expression incorporates both individual learning and coordination, ensuring that agents maximize rewards while fostering cooperation through mutual dependencies. This results in a more cohesive and efficient multi-agent system.

\section{Methodology}

Our research proposes a novel approach, Comparative Advantage Maximization (CAM), designed to enhance individual specialization within multi-agent systems through a two-stage learning framework. This methodology addresses the challenges of heterogeneity in multi-agent systems by first establishing a communal baseline and then fostering individual comparative advantages. By following this structured approach, we ensure that each agent not only starts from a common foundation but also evolves to maximize its unique potential, thus reinforcing the importance of both shared learning and individual specialization.

In Stage 1, we train a population of agents under a centralized conditional policy that maps input states and agent IDs to actions. This policy aims to maximize Mutual Information (MI) across all agent pairs while competing against a mixture of past policies through centralized population self-play. This collective foundation ensures that the agents share a common behavioral understanding, allowing them to build upon this shared experience. Mutual Information, central to this stage, guarantees that agents' behaviors are interlinked, facilitating cooperation and competition \cite{lowe2017multi, foerster2018counterfactual}.

In Stage 2, agents are allowed to individually compete against the base policy from Stage 1, maximizing their Comparative Advantage. This enables agents to specialize beyond the communal policy, leveraging their diverse strengths to outperform the baseline. The transition from collective learning to individual optimization highlights the significance of specialization, reinforcing the notion that while shared learning is essential, it is the divergence into unique strategies that drives true performance improvement.

The following sections provide a detailed explanation of each stage, including the objectives of Mutual Information maximization and Comparative Advantage, as well as how competition is structured within the population learning framework. Each stage builds upon the foundation laid in the previous one, ensuring that the concepts of communal learning and individual specialization reinforce each other in a cyclical process of refinement and optimization.

\subsection{Stage 1: Mutual Information Maximization}

In this stage, we aim to maximize Mutual Information between agents, ensuring that they develop a shared understanding of optimal behaviors while competing in self-play scenarios. We begin by defining a set of distinct agents, denoted as \(N:= \{i, ii, ...\}\), and using a centralized conditional policy \(\{\Pi_{\theta}(a^{id} | s, id)\}_{id= [i, ii, ...]}\) to train agents. The goal is simple yet profound: to maximize mutual information \cite{tishby2000information}, thereby ensuring that the agents' actions and strategies are aligned while retaining enough diversity to allow for individual learning.

Mutual information is computed as:

\begin{equation}
I(X, Y) = \sum_{x, y} p(x, y) \log \left( \frac{p(x, y)}{p(x) p(y)} \right),
\end{equation}

where \(p(x, y)\) is the joint probability of the actions of two agents, and \(p(x)\) and \(p(y)\) are the marginal probabilities of each agent’s actions. By maximizing mutual information, we encourage agents to develop strategies that are coordinated yet distinct, allowing for a shared baseline of behavior that facilitates competition and cooperation.

The agents interact through self-play, refining their strategies by responding to a weighted distribution of past policies. This iterative learning ensures that mutual information is maximized across generations, allowing for both individual and collective improvement \cite{silver2016mastering}. Through this cycle of learning and refinement, agents not only improve their own performance but also contribute to the overall diversity of strategies in the population, reinforcing the notion that diversity leads to robustness.

We maximize the objective function \(J(\theta_{\tau})\), which balances the Q-value returns and decentralized execution, ensuring that agents maximize their mutual information:

\begin{align}
\max J(\theta_{\tau}) &= \sum_{s,a^{i}, a^{ii},id} \left[ Q(s, a^{id}, id) \left[ \Pi_{\theta_{\tau}}(a^{i}, a^{ii} | s, (i, ii)) \log(\Pi_{\theta_{\tau}}(a^{i}, a^{ii} | s, (i, ii))) \right] \right. \nonumber \\
& \quad \left. - V(s, id) \left[ \Pi_{\theta_{\tau}}(a^{i}, a^{ii} | s, (i, ii)) \left[ \log(\Pi_{\theta_{\tau}}(a^{i} | s, i)) + \log(\Pi_{\theta_{\tau}}(a^{ii} | s, ii)) \right] \right] \right].
\end{align}

As the population grows, mutual information generalizes to Interaction Information, ensuring that the communal strategy scales efficiently while maintaining agent diversity \cite{berner2019dota}. This concept of interaction information emphasizes that as more agents interact, the complexity of their relationships increases, leading to richer and more nuanced strategies.

\subsection{Stage 2: Comparative Advantage Maximization}

In Stage 2, we extend the communal learning achieved in Stage 1 by encouraging each agent to specialize and maximize its Comparative Advantage. This transition from collective learning to individual specialization underscores the importance of building on a strong communal foundation while striving for individual excellence. By maximizing comparative advantage, agents can optimize their unique strengths, resulting in strategies that outperform the generalized population.

The Comparative Advantage Maximization of an agent \(x\) can be expressed as the optimization of its policy \(\pi_{\psi^{x}}\) to outperform the communal MIA baseline. This stage is critical, as it allows agents to explore strategies that go beyond what was learned in Stage 1, leveraging their unique attributes to refine their behavior and optimize for performance. The objective function is designed to maximize the $Q$-value of individual actions:

\begin{equation}
\text{Maximize} \ J(\psi^{x}_{\tau}) = \frac{1}{N}\sum^{N}_{id=i} \sum_{(s( \text{MIA}( a^{-id})), a^{x})} \pi_{\psi_{\tau}}(a^{x} | s( \text{MIA}( a^{-id})), id^{x})
\end{equation}

\begin{equation}
\cdot \left[Q(s( \text{MIA}( a^{-id})), a^{x}, id^{x}) \log(\pi_{\psi_{\tau}}(a^{x} | s( \text{MIA}( a^{-id})), id^{x})) \right.
\end{equation}

\begin{equation}
- V( s( \text{MIA}( a^{-id})), id^{x}) \log(\pi_{\psi_{\tau}}(a^{x} | s( \text{MIA}( a^{-id})), id^{x}))\left. \right].
\end{equation}

Agents adapt their policies through competitive training against the fixed MIA policy, progressively refining their strategies to maximize their comparative advantage. As agents specialize, their policies are iteratively added to the opponent pool, creating a cyclical feedback loop of refinement and competition. This continuous cycle of adaptation ensures that agents are constantly evolving, discovering highly specialized strategies that lead to a more robust and efficient multi-agent system \cite{silver2016mastering, lowe2017multi}.

By optimizing for comparative advantage, agents avoid homogenization, developing diverse strategies that lead to a more adaptable and efficient multi-agent system. This diversity, achieved through specialization, not only improves individual performance but also enhances the overall system’s robustness and adaptability \cite{bansal2018emergent, lanctot2017unified}. In this way, Stage 2 of CAM completes the cycle, circling back to the core principles of diversity and specialization that drive the system toward success.
\section{Experiments}

Our research on population-based competitive learning was evaluated using the game \textit{Naruto Mobile}. With over 100 million downloads, this game offers a rich environment to test a wide range of heterogeneous agent behaviors in a fully observable, zero-sum environment. Its real-time mechanics, character diversity, and complex combat strategies make it ideal for assessing how effectively our approach promotes agent specialization. By using this platform, we explore how agents adapt, compete, and specialize in an environment that requires both individual skill and strategic flexibility.

We compared the performance of our 'Specialists' population against NeuPL and Simplex-NeuPL's centralized population self-play approaches. This comparative analysis, based on Competitive Win Rates, Population Size Studies, and Behavioral Diversity, highlights the overgeneralization in existing methods and the improved specialization achieved through our approach. These metrics allow us to demonstrate how CAM fosters both individual optimization and system-wide diversity.

\subsection{Introduction to Naruto Mobile Game Mechanics}

\textit{Naruto Mobile} \cite{liu2023naruto} is a real-time 1 vs 1 game featuring over 300 unique characters, each with its own set of attributes and skills. Before a match, players select a character, an auxiliary skill scroll, and a pet summon. Each character has distinct attributes, such as attack speed, range, and movement capabilities. Characters also possess unique skills with different cooldowns, status effects, ranges, and durations. Common elements across characters include scrolls, pet summons, and a short Invincible skill for evading attacks. The goal of each match is to reduce the opponent’s health points (HP) to zero or have more HP than the opponent when the 60-second match timer ends.

In the context of AI research, these mechanics influence agent strategies, requiring them to adapt to dynamic, real-time decision-making environments \cite{silver2017mastering, bansal2018emergent}. These mechanics force agents to balance offensive and defensive strategies, much like how CAM optimizes the balance between generalization and specialization in agent behavior.

\subsection{Evaluation Metrics}

To evaluate the performance of our method, we employed the following metrics:

\begin{itemize}
    \item \textbf{Competitive Win Rates}: The percentage of matches won by the 'Specialists' population compared to NeuPL and Simplex-NeuPL populations. This metric reflects the effectiveness of specialized strategies in head-to-head competition, aligning with our focus on comparative advantage.
    \item \textbf{Population Size Study}: A comparison of how different population sizes affect performance, especially in terms of scalability and efficiency \cite{bansal2018emergent}. This metric shows how adaptable our method is to varying population sizes.
    \item \textbf{Behavioral Diversity}: A measure of the variety in agent strategies and actions, used to assess the level of specialization and avoid overgeneralization \cite{lanctot2017unified}. By examining diversity, we highlight the importance of specialization and diversity in creating robust, adaptive systems.
\end{itemize}

These metrics provide a comprehensive evaluation of how well CAM promotes both individual excellence and collective adaptability.

\subsection{Results and Analysis}

The 'Specialists' population outperformed across all three metrics. In terms of win rates, Specialists achieved higher victory percentages compared to both NeuPL and Simplex-NeuPL agents. The Population Size Study showed that our method scaled more effectively, with larger populations leading to improved performance, whereas baseline methods experienced diminishing returns. Importantly, the analysis of Behavioral Diversity revealed that the 'Specialists' maintained a greater variety of strategies, avoiding the homogenization often seen in centralized self-play methods.

These results support our hypothesis: maximizing comparative advantage not only improves individual performance but also increases behavioral diversity. By encouraging specialization, we create a population where agents develop unique, adaptive strategies rather than becoming similar to one another. This diversity enhances both the robustness of the system and its ability to adapt to new challenges.

\subsection{Competitive Win Rates}

In this experiment, we evaluated the win rates of the top 8 NeuPL-trained agents and our Specialists. The evaluation was based on N-by-N one-vs-one competitive play within the agent population. Each match pairing consisted of 1,000 games, providing a strong statistical basis for comparison.

\begin{figure}[h]
\centering
\includegraphics[width=0.99\textwidth]{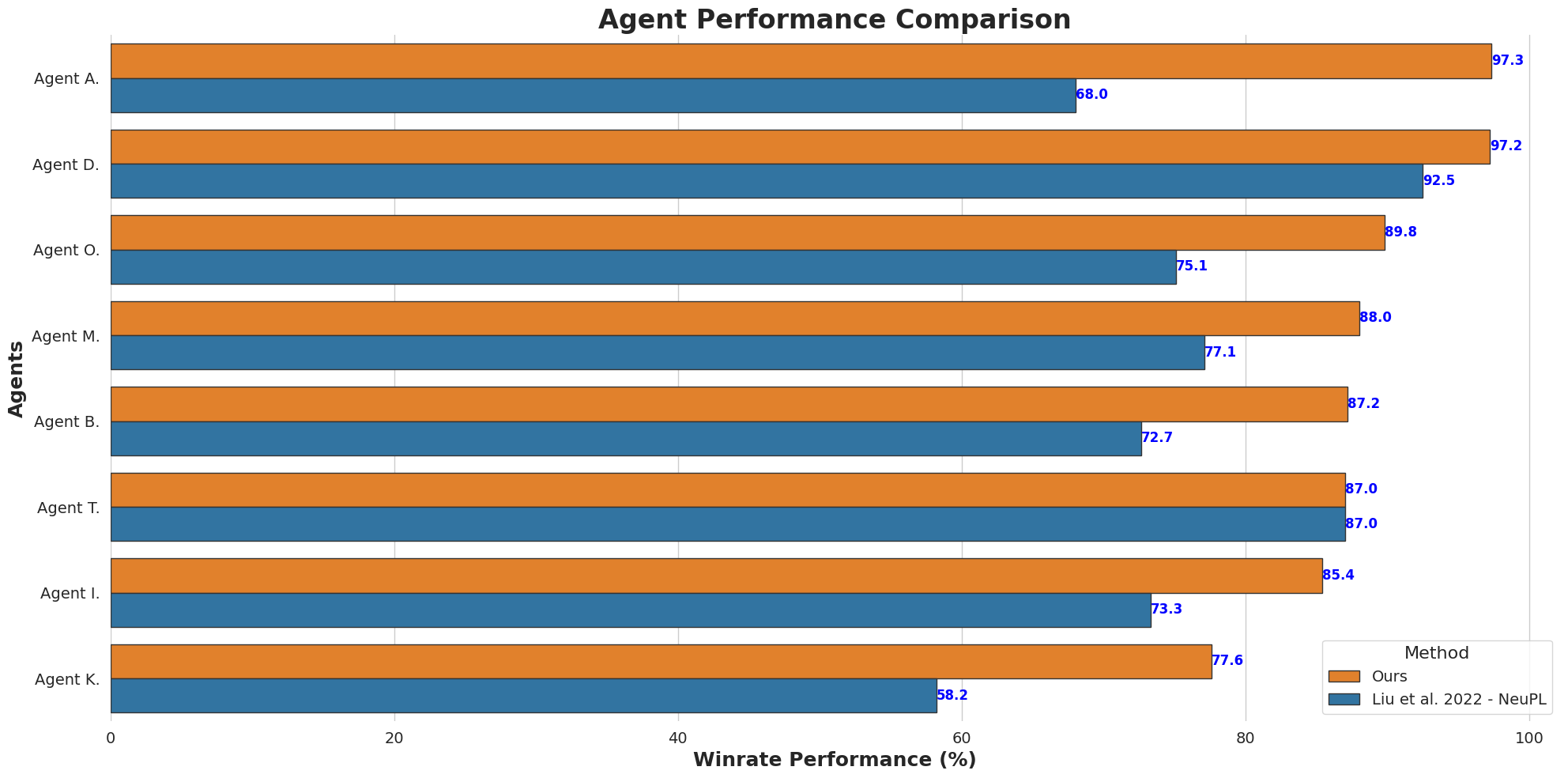}
\caption{\textbf{Expected Win Rates of Specialists and NeuPL}: This figure shows the expected win rates of the top 8 NeuPL agents (blue) and our Specialists (orange) against NeuPL-trained agents. With the exception of Agent T, all Specialists showed improved win rates over their NeuPL counterparts, demonstrating that specialized strategies outperform generalized behaviors.}
\end{figure}

As seen in Figure 1, most agents showed improved win rates after transitioning from NeuPL training to becoming Specialists, with an overall improvement of 13.2\%. This improvement supports the idea that maximizing comparative advantage leads to more specialized, competitive strategies that outperform generalized population behavior \cite{silver2016mastering}.

\subsection{Population Size Study}

In this ablation study, we adjusted the population size for comparative analysis. We evaluated two populations: Population 1, consisting of 8 agents [I, ..., VIII], and Population 2, which expanded to 50 agents. Both populations were first trained using the Simplex-NeuPL method as a baseline \citep{liu2022simplex}, creating two baseline populations: Simplex-8 and Simplex-50. Following this, we applied CAM to enhance specialization, resulting in two new populations: CAM-8 and CAM-50.

The evaluation involved one-vs-one matches across 1,000 games for each match pairing.

\begin{figure}[h]
  \centering
    \includegraphics[width=0.99\textwidth]{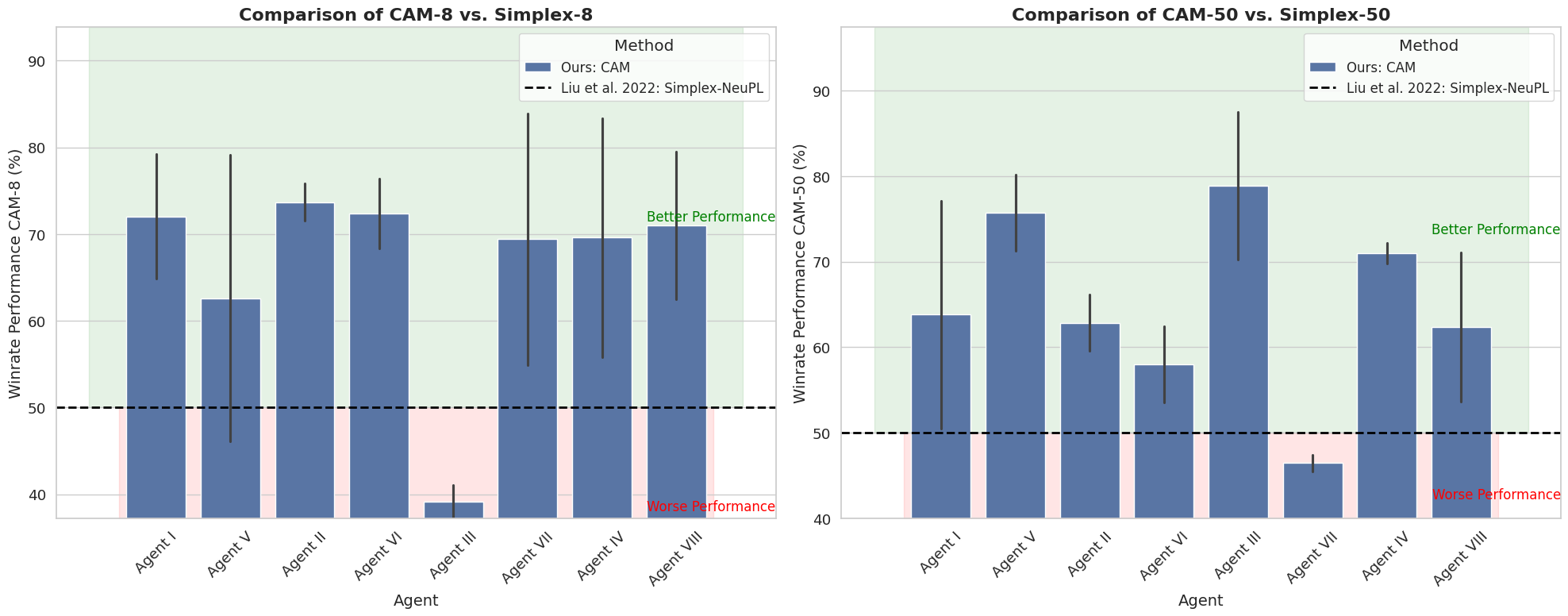}
  \caption{\textbf{Relative Performance of Different Populations}: This figure shows the relative performance of four populations. Green indicates performance improvements from Simplex to CAM, while red indicates a decrease in performance.}
\end{figure}

As shown in Figure 2, 14 out of 16 Specialists outperformed their Simplex counterparts. However, two agents, CAM-8’s Agent III and CAM-50’s Agent VII, exhibited a decrease in performance. This suggests that while most agents benefit from transitioning to specialization, some agents may struggle due to their unique characteristics or interactions with the competitive environment. This highlights that individual agent characteristics must be considered when optimizing for specialization.

Overall, the results demonstrate that CAM enhances specialization and scalability, leading to improved performance in both small and large populations. This transition from generalized strategies to more specialized ones is critical for fostering a system that scales effectively while maintaining individual excellence.

\subsection{Analysis of Behavioral Diversity}

In this experiment, we examined and quantified the behavioral diversity of the agent population before and after applying CAM. By measuring diversity qualitatively and quantitatively, we emphasize that specialization and diversity are both essential for creating robust and adaptive systems.

We tracked the timing and frequency of agent skill usage to gain insights into their behavior.

\begin{figure*}[h]
\includegraphics[width=0.99\textwidth]{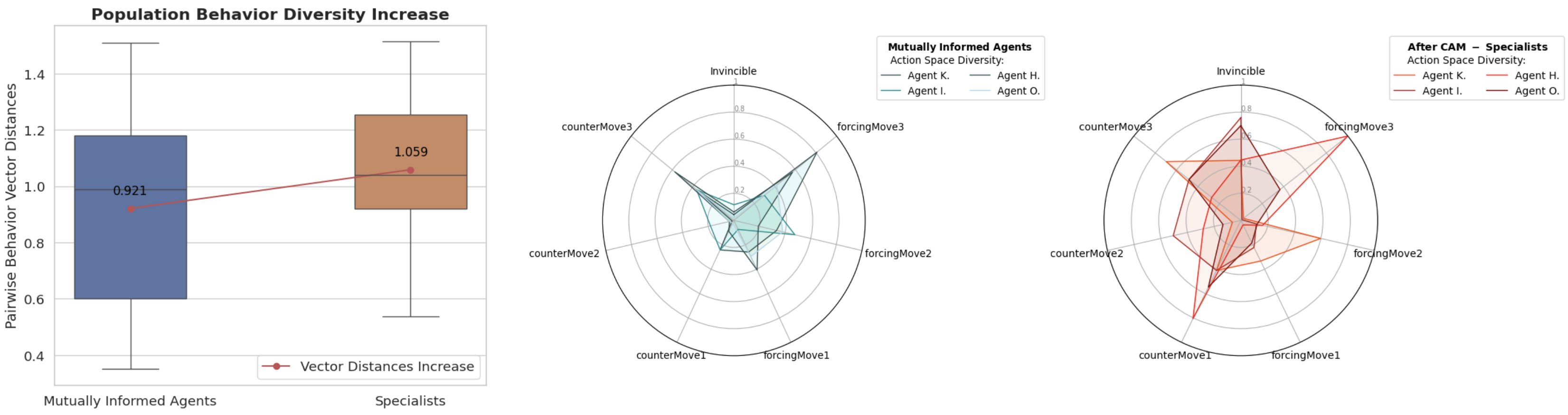}
\caption{\textbf{Assessing Behavioral Diversity}: This figure shows the behavioral diversity of two agent types—\textbf{Mutually Informed Agents} (MIA) and \textbf{Specialists}—in the same character set, facing identical opponents. The radial plots visualize behavioral differences. The distance from the center indicates the frequency of an action, and the direction represents the type of skill used. Skills are categorized into forcingMove (initiating engagement), counterMove (counterattacks), and substitute skill (temporary invincibility). The \textbf{center} of the circle represents the population's mean value, while the \textbf{radial edges} show the maximum deviations.}
\end{figure*}

Figure 3 presents the behavioral strategies of MIA (blue) and Specialists (red). To quantify behavioral diversity, we used action frequency vectors and measured the expected differences in vector distances between agent action pairs. A higher value indicates greater dissimilarity between two agents' actions.

In the left plot of Figure 3, the MIA agents showed an expected vector distance difference of 0.9210, while the Specialists showed a difference of 1.0585—a 14.9\% increase in Euclidean vector distance for the Specialists. This increase in behavioral diversity highlights how specialization fosters unique, adaptive strategies that are more robust in competitive environments.

The distinct variations in the red radial plot suggest that Specialist agents employ more diverse strategies, while the MIA agents (in blue) display more uniform behavior. This analysis shows that CAM enhances the behavioral diversity of agents, making them more adaptive and specialized \citep{bansal2018emergent, lanctot2017unified}. Once again, we see how CAM’s emphasis on specialization and diversity leads to superior outcomes in multi-agent learning.
\section{Conclusion}

In this study, we investigated the potential of individual specialization in Multi-Agent Reinforcement Learning (MARL) by introducing Comparative Advantage Maximization (CAM), a new approach designed to promote individual excellence in competitive environments. CAM moves beyond generalization and emphasizes the critical role of specialization, highlighting that agents must develop unique strengths to succeed in complex settings. Our two-stage framework—first maximizing Mutual Information and then focusing on Comparative Advantage—fosters both collective learning and individual specialization. This cyclic relationship between shared knowledge and personal optimization is central to our approach, ensuring that agents grow not only in coordination but also in diversity.

Our experiments in the complex environment of \textit{Naruto Mobile} demonstrated the significant performance benefits of our method. The CAM-trained populations consistently outperformed their counterparts, achieving a 13.2\% increase in individual victory rates and a 14.9\% increase in behavioral diversity across different population setups. These results emphasize that specialization, when built on a foundation of shared learning, drives competitive success. Victory rates and behavioral diversity—both crucial for evaluating agent performance—support the key principle that maximizing comparative advantage leads to better outcomes.

The findings suggest that transitioning from generalization to specialization enables agents to leverage their unique characteristics, resulting in more efficient learning in MARL. This conclusion reinforces our central thesis: that fostering diversity in strategies not only prevents overgeneralization but also unlocks new paths to optimization. Specialization allows agents to capitalize on their distinct advantages, enhancing the system's adaptability and robustness. This demonstrates that individual specialization plays a crucial role in boosting performance, showing that promoting diversity in agent strategies leads to better results \cite{bansal2018emergent, lanctot2017unified}. The results consistently confirm that diversity, rooted in specialization, is a key factor for success in multi-agent systems.

However, one limitation of our approach is the additional computational cost involved in learning individualized policies for each agent in Stage 2. While this process enables agents to specialize, the time required to achieve this can become considerable for larger populations. This trade-off between specialization and computational efficiency is a challenge, reflecting the broader tension between individual optimization and system-wide efficiency. Future research could explore clustering the agent population into smaller groups to reduce computational costs while maintaining some level of specialization \cite{silver2016mastering}. By clustering, we may be able to balance efficiency and diversity, preserving the benefits of specialization while minimizing the computational burden.

In conclusion, our research makes significant contributions to MARL by showcasing the advantages of individual specialization. Specialization, as achieved through CAM, not only improves individual agent performance but also enhances the collective intelligence of the system. Our findings suggest that incorporating specialization into MARL algorithms can lead to more efficient learning and better overall agent performance. These insights, which resonate throughout our study, reiterate the core message: specialization leads to excellence. We believe that this work opens new avenues for future research in MARL and related fields, and we are optimistic about the innovations that this research will inspire. As future research builds on this foundation, we expect specialization to become an even more central theme in multi-agent systems, driving both theoretical progress and practical applications forward.

\bibliography{iclr2024_conference}

\begin{thebibliography}{24}
\providecommand{\natexlab}[1]{#1}
\providecommand{\url}[1]{\texttt{#1}}
\expandafter\ifx\csname urlstyle\endcsname\relax
  \providecommand{\doi}[1]{doi: #1}\else
  \providecommand{\doi}{doi: \begingroup \urlstyle{rm}\Url}\fi

\bibitem[Abadi et~al.(2015)Abadi, Agarwal, Barham, Brevdo, Chen, Citro, Corrado, Davis, Dean, Devin, Ghemawat, Goodfellow, Harp, Irving, Isard, Jia, Jozefowicz, Kaiser, Kudlur, Levenberg, Man\'{e}, Monga, Moore, Murray, Olah, Schuster, Shlens, Steiner, Sutskever, Talwar, Tucker, Vanhoucke, Vasudevan, Vi\'{e}gas, Vinyals, Warden, Wattenberg, Wicke, Yu, and Zheng]{tensorflow2015-whitepaper}
Mart\'{i}n Abadi, Ashish Agarwal, Paul Barham, Eugene Brevdo, Zhifeng Chen, Craig Citro, Greg~S. Corrado, Andy Davis, Jeffrey Dean, Matthieu Devin, Sanjay Ghemawat, Ian Goodfellow, Andrew Harp, Geoffrey Irving, Michael Isard, Yangqing Jia, Rafal Jozefowicz, Lukasz Kaiser, Manjunath Kudlur, Josh Levenberg, Dandelion Man\'{e}, Rajat Monga, Sherry Moore, Derek Murray, Chris Olah, Mike Schuster, Jonathon Shlens, Benoit Steiner, Ilya Sutskever, Kunal Talwar, Paul Tucker, Vincent Vanhoucke, Vijay Vasudevan, Fernanda Vi\'{e}gas, Oriol Vinyals, Pete Warden, Martin Wattenberg, Martin Wicke, Yuan Yu, and Xiaoqiang Zheng.
\newblock {TensorFlow}: Large-scale machine learning on heterogeneous systems, 2015.
\newblock URL \url{https://www.tensorflow.org/}.
\newblock Software available from tensorflow.org.

\bibitem[Bakhtin et~al.(2022)Bakhtin, Muntean, Meier, Adam, Bordes, Weston, and Szlam]{bakhtin2022neupl}
Anton Bakhtin, Petre Muntean, Franziska Meier, Hartmut Adam, Antoine Bordes, Jason Weston, and Arthur Szlam.
\newblock Neupl: Neural population-based learning.
\newblock \emph{arXiv preprint arXiv:2210.07377}, 2022.

\bibitem[Bansal et~al.(2018)Bansal, Pachocki, Sidor, Sutskever, and Mordatch]{bansal2018emergent}
Trapit Bansal, Jakub Pachocki, Szymon Sidor, Ilya Sutskever, and Igor Mordatch.
\newblock Emergent complexity via multi-agent competition.
\newblock In \emph{Proceedings of the 6th International Conference on Learning Representations}, 2018.

\bibitem[Berner et~al.(2019)Berner, Brockman, Chan, Cheung, Debiak, Dennison, Farhi, Fischer, Hashme, Hesse, et~al.]{berner2019dota}
Christopher Berner, Greg Brockman, Brooke Chan, Vicki Cheung, Przemek Debiak, Christy Dennison, David Farhi, Quirin Fischer, Shariq Hashme, Christopher Hesse, et~al.
\newblock Dota 2 with large scale deep reinforcement learning.
\newblock \emph{arXiv preprint arXiv:1912.06680}, 2019.

\bibitem[Espeholt et~al.(2018)Espeholt, Soyer, Munos, Simonyan, Mnih, Ward, Doron, Firoiu, Harley, Dunning, et~al.]{espeholt2018impala}
Lasse Espeholt, Hubert Soyer, Remi Munos, Karen Simonyan, Vlad Mnih, Tom Ward, Yotam Doron, Vlad Firoiu, Tim Harley, Iain Dunning, et~al.
\newblock Impala: Scalable distributed deep-rl with importance weighted actor-learner architectures.
\newblock In \emph{International Conference on Machine Learning}, pp.\  1407--1416. PMLR, 2018.

\bibitem[Foerster et~al.(2018)Foerster, Farquhar, Afouras, Nardelli, and Whiteson]{foerster2018counterfactual}
Jakob Foerster, Gregory Farquhar, Triantafyllos Afouras, Nantas Nardelli, and Shimon Whiteson.
\newblock Counterfactual multi-agent policy gradients.
\newblock In \emph{Proceedings of the AAAI Conference on Artificial Intelligence}, volume~32, 2018.

\bibitem[Lanctot et~al.(2017)Lanctot, Zambaldi, Gruslys, Lazaridou, Tuyls, P{\'e}rolat, Silver, and Graepel]{lanctot2017unified}
Marc Lanctot, Vinicius Zambaldi, Audrunas Gruslys, Angeliki Lazaridou, Karl Tuyls, Julien P{\'e}rolat, David Silver, and Thore Graepel.
\newblock A unified game-theoretic approach to multiagent reinforcement learning.
\newblock \emph{Advances in neural information processing systems}, 30, 2017.

\bibitem[Littman(1994)]{littman1994markov}
Michael~L Littman.
\newblock Markov games as a framework for multi-agent reinforcement learning.
\newblock In \emph{Proceedings of the eleventh international conference on machine learning}, pp.\  157--163. Citeseer, 1994.

\bibitem[Liu et~al.(2023)Liu, Wang, and Zhang]{liu2023naruto}
J.~Liu, Y.~Wang, and Q.~Zhang.
\newblock Naruto mobile: Competitive multi-agent dynamics in real-time decision-making environments.
\newblock \emph{AI Game Studies Journal}, 2023.

\bibitem[Liu et~al.(2022{\natexlab{a}})Liu, Lanctot, Marris, and Heess]{liu2022simplex}
Siqi Liu, Marc Lanctot, Luke Marris, and Nicolas Heess.
\newblock Simplex neural population learning: Any-mixture bayes-optimality in symmetric zero-sum games.
\newblock In \emph{International Conference on Machine Learning}, pp.\  13793--13806. PMLR, 2022{\natexlab{a}}.

\bibitem[Liu et~al.(2022{\natexlab{b}})Liu, Marris, Hennes, Merel, Heess, and Graepel]{liu2022neupl}
Siqi Liu, Luke Marris, Daniel Hennes, Josh Merel, Nicolas Heess, and Thore Graepel.
\newblock Neupl: Neural population learning.
\newblock \emph{arXiv preprint arXiv:2202.07415}, 2022{\natexlab{b}}.

\bibitem[Lowe et~al.(2017)Lowe, Wu, Tamar, Harb, Abbeel, and Mordatch]{lowe2017multi}
Ryan Lowe, Yi~Wu, Aviv Tamar, Jean Harb, Pieter Abbeel, and Igor Mordatch.
\newblock Multi-agent actor-critic for mixed cooperative-competitive environments.
\newblock In \emph{Advances in neural information processing systems}, pp.\  6379--6390, 2017.

\bibitem[Mnih et~al.(2015)Mnih, Kavukcuoglu, Silver, Rusu, Veness, Bellemare, Graves, Riedmiller, Fidjeland, Ostrovski, et~al.]{mnih2015human}
Volodymyr Mnih, Koray Kavukcuoglu, David Silver, Andrei~A Rusu, Joel Veness, Marc~G Bellemare, Alex Graves, Martin Riedmiller, Andreas~K Fidjeland, Georg Ostrovski, et~al.
\newblock Human-level control through deep reinforcement learning.
\newblock \emph{Nature}, 518\penalty0 (7540):\penalty0 529--533, 2015.

\bibitem[Rashid et~al.(2018)Rashid, Samvelyan, De~Witt, Farquhar, Foerster, and Whiteson]{rashid2018qmix}
Tabish Rashid, Mikayel Samvelyan, Christian~S De~Witt, Gregory Farquhar, Jakob Foerster, and Shimon Whiteson.
\newblock Qmix: Monotonic value function factorization for deep multi-agent reinforcement learning.
\newblock In \emph{International Conference on Machine Learning}, pp.\  4295--4304, 2018.

\bibitem[Russell \& Norvig(1995)Russell and Norvig]{russell1995artificial}
Stuart Russell and Peter Norvig.
\newblock \emph{Artificial intelligence: a modern approach}.
\newblock Prentice Hall, 1995.

\bibitem[Schulman et~al.(2017)Schulman, Wolski, Dhariwal, Radford, and Klimov]{schulman2017proximal}
John Schulman, Filip Wolski, Prafulla Dhariwal, Alec Radford, and Oleg Klimov.
\newblock Proximal policy optimization algorithms.
\newblock \emph{arXiv preprint arXiv:1707.06347}, 2017.

\bibitem[Sergeev \& Balso(2018)Sergeev and Balso]{sergeev2018horovod}
Alexander Sergeev and Mike~Del Balso.
\newblock Horovod: fast and easy distributed deep learning in {TensorFlow}.
\newblock \emph{arXiv preprint arXiv:1802.05799}, 2018.

\bibitem[Silver et~al.(2016)Silver, Huang, Maddison, Guez, Sifre, Van Den~Driessche, Schrittwieser, Antonoglou, Panneershelvam, Lanctot, et~al.]{silver2016mastering}
David Silver, Aja Huang, Chris~J Maddison, Arthur Guez, Laurent Sifre, George Van Den~Driessche, Julian Schrittwieser, Ioannis Antonoglou, Veda Panneershelvam, Marc Lanctot, et~al.
\newblock Mastering the game of go with deep neural networks and tree search.
\newblock \emph{nature}, 529\penalty0 (7587):\penalty0 484--489, 2016.

\bibitem[Silver et~al.(2017)Silver, Schrittwieser, Simonyan, Antonoglou, Huang, Guez, Hubert, Baker, Lai, Bolton, et~al.]{silver2017mastering}
David Silver, Julian Schrittwieser, Karen Simonyan, Ioannis Antonoglou, Aja Huang, Arthur Guez, Thomas Hubert, Lucas Baker, Matthew Lai, Adrian Bolton, et~al.
\newblock Mastering the game of go without human knowledge.
\newblock \emph{nature}, 550\penalty0 (7676):\penalty0 354--359, 2017.

\bibitem[Stone \& Veloso(2000)Stone and Veloso]{stone2000multiagent}
Peter Stone and Manuela Veloso.
\newblock Multiagent systems: A survey from a machine learning perspective.
\newblock \emph{Autonomous Robots}, 8\penalty0 (3):\penalty0 345--383, 2000.

\bibitem[Sutton \& Barto(2018)Sutton and Barto]{sutton2018reinforcement}
Richard~S Sutton and Andrew~G Barto.
\newblock \emph{Reinforcement learning: An introduction}.
\newblock MIT press, 2018.

\bibitem[Tishby et~al.(2000)Tishby, Pereira, and Bialek]{tishby2000information}
Naftali Tishby, Fernando~C Pereira, and William Bialek.
\newblock The information bottleneck method.
\newblock In \emph{Proceedings of the 37th annual Allerton conference on communication, control and computing}, pp.\  368--377, 2000.

\bibitem[Wang et~al.(2020)Wang, Zhu, Geng, Zhang, Tang, and Qin]{wang2020adaptive}
Tao Wang, Yizhou Zhu, Yu~Geng, Wei Zhang, Zhiming Tang, and Tao Qin.
\newblock Adaptive agent coordination in multi-agent reinforcement learning via learned communication.
\newblock In \emph{Proceedings of the 34th AAAI Conference on Artificial Intelligence}, pp.\  7646--7653, 2020.

\bibitem[Watkins \& Dayan(1992)Watkins and Dayan]{watkins1992q}
Christopher~JCH Watkins and Peter Dayan.
\newblock Q-learning.
\newblock \emph{Machine learning}, 8\penalty0 (3-4):\penalty0 279--292, 1992.

\end{thebibliography}
\bibliographystyle{iclr2024_conference}
\clearpage
\section{Appendix}

\subsection{Architecture of the Mobile Deep Learning Model}

\begin{figure*}[ht]
\vskip 0.2in
\centering
\includegraphics[width=0.99\textwidth]{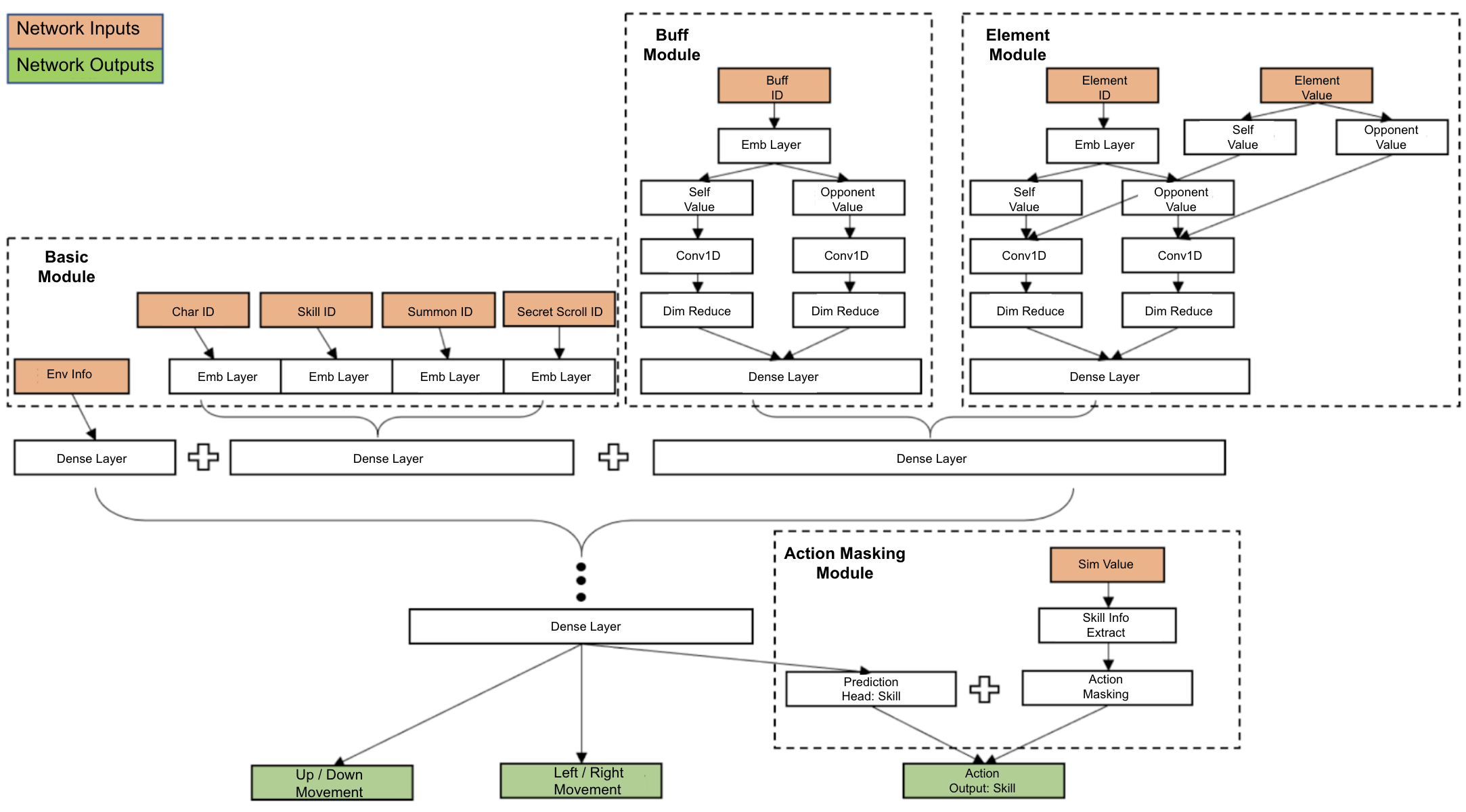}
\caption{\textbf{Deep Reinforcement Learning (DRL) Architecture for Mobile Devices:} Our neural network, deployable on mobile devices, employs an embedding layer to reduce the dimensionality of the inputs. To minimize computational load, we use Conv1D layers instead of Dense layers. The network is divided into four modules from input to output.}
\vskip -0.2in
\end{figure*}

The four modules are as follows:

\begin{enumerate}
    \item \textbf{Basic Module:} Processes game environment statistics and character-specific information.
    \item \textbf{Buff Module:} Handles buff and debuff information during gameplay.
    \item \textbf{Element Module:} Manages characters' equipment, weapon IDs, scroll IDs, etc.
    \item \textbf{Action Masking Module:} Provides the model with information on the availability of each action.
\end{enumerate}

The modular structure depicted in Figure 4 allows for sparse connectivity within the model, reducing computational load compared to fully connected layers. This efficient structure enables local inference on mobile devices with limited computational resources.

\textbf{Markov Decision Process (MDP):} The model receives input states, $s_t$, from the Basic, Buff, and Element Modules. These states help identify characters, equipped Summons and Scrolls, characters' positions, movements, and skills' buff and element information. The model then predicts an action output, $a_t$, to control the 2D movement of the agent and the available attack and skills. The reward can be customized, but in our standard mechanics, it is based on the weighted sum of an agent's own HP(10), opponent's HP(10), the result of the battle(10), combo(5), and mana(5). With the transition of the action, new state $s_{t+1}$ is given to the model for the next iteration of MDP. \par

\subsection{Hyperparameters and Hardware Used}
\textbf{Hyperparameters}
\label{hyp}
\begin{itemize}
    \item PPO: 0.1
    \item n-step: 100 frames
    \item Reward discount factor: 0.995
    \item Learning rate: 1e-4
\end{itemize}
\textbf{Hardware Used}
\begin{itemize}
    \item CPUS: 5,300
    \item GPUS: 0
    \item Batch size: 300
    \item Compute Time:
    \begin{itemize}
        \item $\approx$ 180 Hrs
    \end{itemize}
\end{itemize}

\subsection{Algorithm Pseudocode}\label{code}

In this section, we present the pseudocode for our Comparative Advantage Maximization (CAM) specialists learning approach.

The CAM method optimizes individual agent policies within a heterogeneous population $\mathcal{P} = \{1, 2, \dots, N\}$, given a pre-trained Mutually Informed Agents (MIA) policy $\Pi_{\theta_*}$. The goal is to maximize the comparative advantage of each individual agent's policy against the MIA policies $\Pi_{\theta_*}$. Matchmaking for each one-vs-one scenario is prioritized based on the interaction graph updated by the graph solver $F$. After a batch of episodes using Neural Population Learning (NPL), each agent's policy is optimized with Proximal Policy Optimization (PPO).

\begin{algorithm}
\caption{CAM Multi-Agent Specialization Algorithm}
\label{alg:CAMMAS}
\SetKwInOut{Input}{Input}
\SetKwInOut{Parameter}{Parameters}
\SetKwInOut{Output}{Output}

\Input{
    Population of agents $\mathcal{P} = \{1, 2, \dots, N\}$
}
\Parameter{
    Pre-trained MIA policy $\Pi_{\theta_*}$\;
    Agent-specific initial policies $\{\pi_{\psi_n}^0\}_{n=1}^N$ initialized from $\Pi_{\theta_*}$
}
\Output{
    Optimized individual policies $\{\pi_{\psi_n}^T\}_{n=1}^N$
}

\BlankLine
Initialize iteration counter $\tau \leftarrow 0$\;

\While{not converged}{
    \ForEach{agent $n \in \mathcal{P}$}{
        \tcp{Construct agent $n$'s interaction graph}
        $\Sigma_n^\tau \leftarrow F(U_n^{\tau-1})$\;
        
        \tcp{Select opponent policies based on interaction graph}
        $\Pi_{\Sigma_n^\tau} \leftarrow \{\pi_{\psi_i}^\tau \mid i \in \Sigma_n^\tau\}$\;
        
        \tcp{Collect experience via NPL}
        $\mathcal{D}_n^\tau \leftarrow$ NPL($\pi_{\psi_n}^\tau$, $\sigma_n$, $\Pi_{\Sigma_n^\tau}$, $T$)\;
        
        \tcp{Update agent $n$'s policy using PPO}
        $\pi_{\psi_n}^{\tau+1} \leftarrow$ PPO($\pi_{\psi_n}^\tau$, $\mathcal{D}_n^\tau$)\;
        
        \tcp{Evaluate agent $n$'s performance}
        $U_n^\tau \leftarrow$ Evaluate($\pi_{\psi_n}^{\tau+1}$, $\Pi^N$)\;
    }
    \tcp{Update the opponent pool with the latest policies}
    $\Pi^N \leftarrow \{\pi_{\psi_n}^{\tau+1}\}_{n=1}^N$\;
    
    Increment iteration counter $\tau \leftarrow \tau + 1$\;
}
\end{algorithm}

In Algorithm \ref{alg:CAMMAS}, $U_n^\tau$ represents the interaction graph for agent $n$, capturing the probabilistic outcomes of all pairwise matches involving agent $n$. The graph solver $F(U_n^\tau)$ updates the interaction graph by prioritizing adversarial opponents for agent $n$. This process iterates until the performance of the population converges.

The NPL component simulates matches between agents' policies, where each agent's policy is matched against selected opponent policies from the MIA population or other agents. Interaction trajectories are stored in a replay buffer for policy gradient optimization.

\begin{algorithm}
\SetAlgoLined
\caption{Neural Population Learning (NPL)}
\label{alg:NPL}
\SetKwInOut{Input}{Input}
\SetKwInOut{Output}{Output}

\Input{
    Agent policy $\pi$\;
    Opponent selection strategy $\sigma$\;
    Opponent policy set $\Pi$\;
    Number of episodes $T$
}
\Output{
    Replay buffer $\mathcal{D}$ containing collected trajectories
}

\BlankLine
Initialize replay buffer $\mathcal{D} \leftarrow \emptyset$\;
\For{$\text{episode} = 1$ \KwTo $T$}{
    \tcp{Select opponent policy using strategy $\sigma$}
    $\pi_{\text{opp}} \leftarrow \sigma(\Pi)$\;
    
    \tcp{Initialize environment with policies $(\pi, \pi_{\text{opp}})$}
    Initialize environment $\mathcal{E}$ with agents using $\pi$ and $\pi_{\text{opp}}$\;
    
    \tcp{Collect trajectory from interaction}
    $\tau \leftarrow$ RunEpisode($\mathcal{E}$)\;
    
    \tcp{Update replay buffer}
    $\mathcal{D} \leftarrow \mathcal{D} \cup \{\tau\}$\;
}
\Return{$\mathcal{D}$}
\end{algorithm}

\subsection{Development of the Centralized Population Self-play Algorithm for Mutually Informed Agents}
Our study employs the multi-agent interaction graph of NeuPL \citep{liu2022neupl} for neural population learning in the conditional population net. The nodes in this graph represent different generations of agents and are interconnected by weighted edges denoted by $\Sigma^{(x,y)} \in \mathbb{R}^{N \times N}$. NeuPL provides a population self-play framework that not only pits the current population $\Pi^{\text{id}}{\theta{\tau}}$ against a variety of distinct agents but also assigns priority to the weighted edges for different generations of the $\epsilon$-NE population $\Pi^{\text{id}}{\theta{0:\tau-1}}$. Each population is depicted as a conditional population net that acquires a set of best-response (BR) strategies against all previous generations of multi-agent mixed strategies.
\begin{figure}[h]
\vskip 0.2in
\centering
\includegraphics[width=0.8\textwidth]{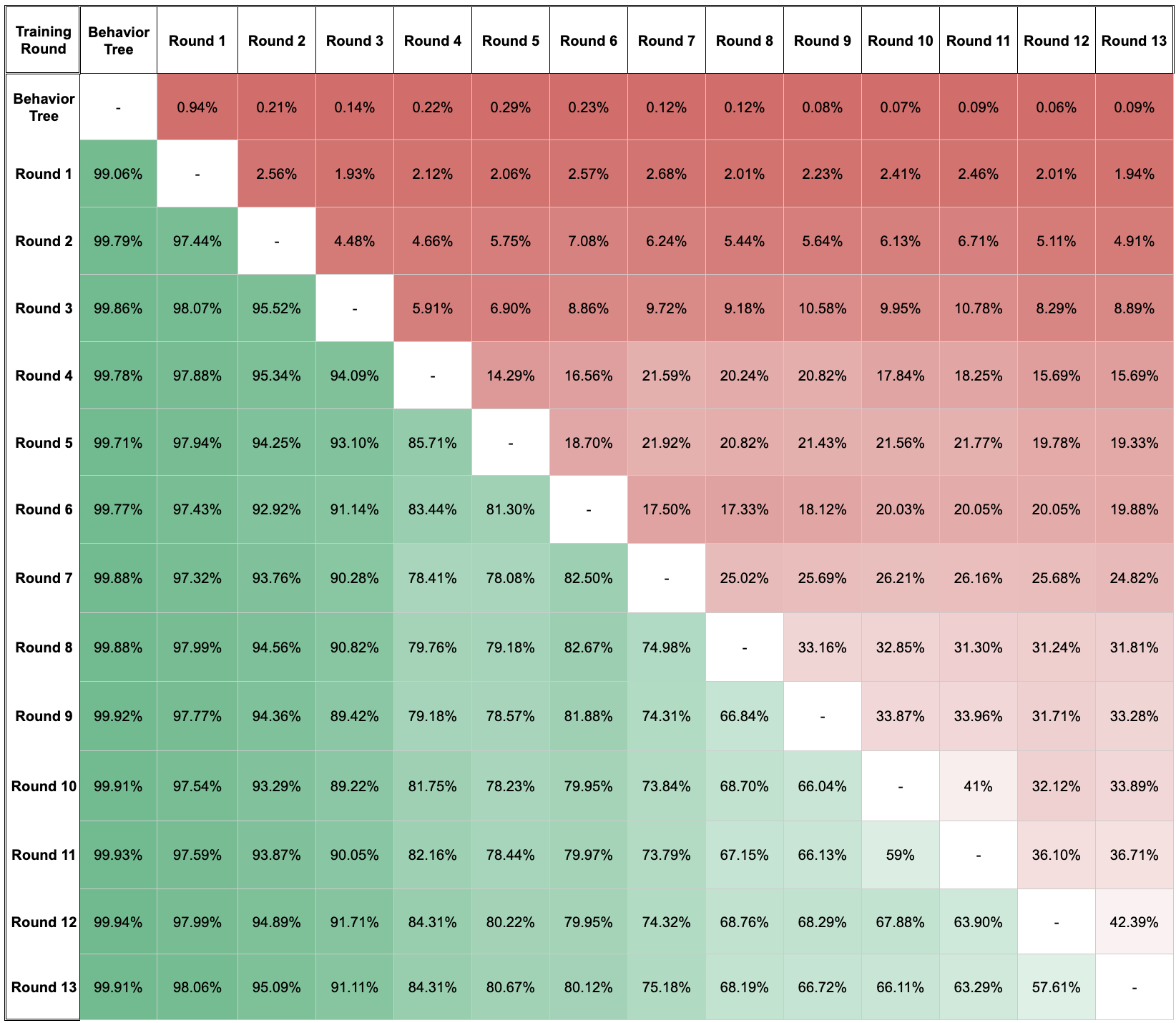}
\caption{Evaluation matches across different iterations of \textbf{Mutually Informed Agents (MIA)} $\Pi^{id}{\theta_{t}}$. The heatmap illustrates a decrease in MIA's collective learning as the training approaches the 11th to 13th iteration. Notably, the 13th iteration has a win rate of merely 57.6\% against the 12th iteration of MIA, which is proximate to the Nash equilibrium of 50\%.}
\vskip -0.2in
\end{figure}

We conduct an evaluation of all rounds of the MIA policies against each other. As depicted in Figure 5, the heatmap demonstrates the monotonic convergence of the MIA population. By the 13th iteration, the MIA population has converged to an $\varepsilon -$ Nash Equilibrium, where $\varepsilon \approx 7.6\%$. Further training might marginally enhance the performance, but it risks training collapse.

\subsection{Software Utilization and Licensing}
The models employed in our study are constructed using a variety of software libraries, which include TensorFlow, TensorFlow Lite \citep{tensorflow2015-whitepaper}, IMPALA \citep{espeholt2018impala}, and Horovod \citep{sergeev2018horovod}. 

\textbf{TensorFlow} was utilized to build and train the deep reinforcement learning models, while \textbf{TensorFlow Lite} enabled deployment on mobile devices. \textbf{IMPALA}, a distributed reinforcement learning framework, was used to ensure scalability in training, and \textbf{Horovod} facilitated distributed training across multiple CPUs.

These software libraries are all licensed under the \textbf{Apache License 2.0}. This open-source license permits the use, modification, and distribution of the software, provided that:
\begin{itemize}
    \item A copy of the license is included with any distribution of the software.
    \item Any modifications or derivative works that are distributed must also fall under the Apache License 2.0.
\end{itemize}

For more information about the terms and conditions of this license, please refer to the official \href{https://www.apache.org/licenses/LICENSE-2.0}{Apache License 2.0} documentation.

It is important to note that any derivative works created using these libraries in our research will also be subject to the terms of the Apache License 2.0, ensuring compliance with open-source licensing obligations.

\subsection{Acknowledgement}
This research was made possible through the financial support of Tencent, in collaboration with The Hong Kong University of Science and Technology (HKUST). We would like to express our sincere gratitude to the following individuals and teams for their invaluable contributions:

\begin{itemize}
    \item \textbf{AI Development:} Zengw Zeng, Julie Wang, Gustav Zhang, Vino Wan, Caron Zhang, Wind Li, Town Zhang, for their expertise and dedication throughout the research process.
    \item \textbf{Graphics Design:} Rickie Luo, Sylvie Yang, Kun Zheng, Leo Q. Li, Irrion Zhao, Z. H. Zhao, X. H. Zhu, Zhiqian Yao, Nicole Xin Yu, Miffy Wu, Shuai GH, Eien Song, Xie Jiazhang, Dijing Zhao, and Fishny Wang, for their outstanding work in creating the visual elements.
    \item \textbf{Development Production:} Peck Huang, Fineman Xie, Jump Chen, Jerry Qin, Leo Z. Zhang, Logan, Lgnatz Jiang, Z. Yuan, and L. Zhu, for their commitment and hard work in the project’s development.
    \item \textbf{Risk Management:} K. J. K. S. Chen, Saman Zhang, Gerhard He, X. Zhuang, S. Huang, and C. Fang, for their critical support in risk evaluation and management.
    \item \textbf{Sound Production:} Olivier Ma, Lazy Huang, and Y. Yong, for their contributions to the audio design and production.
    \item \textbf{User Experience Design:} Moson Lin, Claren Xu, for their efforts in enhancing the user experience.
    \item \textbf{Strategic Planning:} Loyn Liu, Yuanzhen Li, Faker Liu, Dido Sun, Roy Gao, Yongshu Hu, Tracy M. Qi, Small Cai, and Jiantong Li, for their strategic oversight and planning.
    \item \textbf{Operations:} Sean Shen, Chi Ding, C. K. Liao, and A. Zhang, for their operational support and coordination throughout the project.
    \item \textbf{Inspection and Supervision:} Tokishi and Shane Cui, for their diligence in overseeing project inspections and quality control.
    \item \textbf{Anticheating:} Kero Cao, for ensuring the integrity of the project through effective anti-cheating measures.
\end{itemize}

We also extend our gratitude to our colleagues at Tencent IEG and Cros for their collective expertise and continuous support, which were crucial to the success of this project.

\end{document}